\documentclass[runningheads]{llncs}

\usepackage{eccv}

\usepackage{eccvabbrv}

\usepackage{graphicx}
\usepackage{booktabs}
\usepackage{esvect}

\usepackage[accsupp]{axessibility}  

\usepackage{algorithm}
\usepackage{algpseudocodex}
\usepackage{booktabs}
\usepackage{multirow}
\usepackage{array}
\usepackage{amsmath}
\usepackage{wrapfig}
\usepackage{nicefrac}
\usepackage{caption}
\usepackage{subcaption}
\usepackage[dvipsnames]{xcolor}
\usepackage{rotating}

\newcommand{\METHODNAME}{GATO-Vid}

\newcommand{\PreserveBackslash}[1]{\let\temp=\\#1\let\\=\temp}
\newcolumntype{C}[1]{>{\PreserveBackslash\centering}p{#1}}
\newcolumntype{R}[1]{>{\PreserveBackslash\raggedleft}p{#1}}
\newcolumntype{L}[1]{>{\PreserveBackslash\raggedright}p{#1}}
\allowdisplaybreaks

\newif\ifcomments
\commentsfalse

\commentsfalse
\commentstrue

\usepackage{hyperref}

\usepackage{orcidlink}

\begin{document}

\title{Spatially-Grounded Text-to-Video Generation via Inference-Time Gradient-Free Optimization} 

\titlerunning{\METHODNAME}

\author{Guillaume Jeanneret\inst{1}\orcidlink{0000-0002-0055-7816}
\and
Mathis Koroglu\inst{1,2}\orcidlink{0009-0000-6589-3151}
\and
Hugo Caselles-Dupr\'e\inst{2}\orcidlink{0000-0003-2711-5732}
\and
Arnaud Dapogny\inst{1}
\and
Matthieu Cord\inst{1,3}\orcidlink{0000-0002-0627-5844}
}

\authorrunning{G.~Jeanneret \etal}

\institute{Sorbonne Université, CNRS, ISIR, Paris, France \and
Obvious Research, Paris, France \and
Valeo.ai, Paris, France}

\maketitle

\begin{abstract}
Diffusion Transformer Text-to-Video models have achieved remarkable synthesis quality, yet fine-grained spatial controllability remains a significant challenge. While existing training-free methods produce solid overall results in spatially grounded generation, \ie, placing a specific object in a designated location, they rely on gradient-based optimization techniques that incur prohibitive computational overhead, a bottleneck amplified in modern large-scale architectures. To address this limitation, we present Gradient-free Analytical Trajectory Optimization Video Generation (\METHODNAME), a novel training-free and gradient-free approach for precise spatial guidance. Rather than relying on costly backward passes, we introduce an alternative cross-attention score and solve it analytically to obtain an exact, closed-form solution. To use our analytical solution, we propose an on-the-fly injection mechanism tailored to the topological manifold of the transformer's latent space. Our experiments demonstrate that \METHODNAME{} significantly outperforms existing baselines in localization accuracy while introducing minimal computational overhead.
\end{abstract}

\section{Introduction}


Text-to-Video (T2V) diffusion models have made significant progress, producing high-quality videos that are coherent, realistic, and well-aligned with textual prompts. These improvements have been driven by advances in Diffusion Transformer (DiT) architectures~\cite{Peebles_2023_ICCV}, large-scale video-text training data, and the scaling of generative models to billions of parameters~\cite{Wang2025WanOA,yang2025cogvideox}. However, text remains a limited interface for controlling video generation. A prompt can describe the presence of objects, their appearance, and their general motion, but cannot reliably specify spatio-temporal localization.

To go beyond this controllability limitation, we address the task of spatially localized text-to-video generation. Given a text prompt and spatial constraints, such as bounding boxes, the goal of this a task is to generate a video that both faithfully follows the prompt and respects the prescribed object locations. In the literature, existing approaches can be divided into training-based and training-free methods. Training-based approaches learn spatial control from annotated data~\cite{Feng_2025_CVPR}, either by training dedicated modules or by fine-tuning existing generative models. However, this comes at a significant cost, requiring substantial computational resources and large-scale annotated data given the scale of recent text-to-video models. For this reason, training-free methods are attractive alternatives as they only control pretrained generators at inference time, without modifying their weights or collecting new data. Despite that, current training-free methods also face limitations. The most effective approaches often rely on gradient-based guidance~\cite{mao2025zerotrail}, which requires backpropagating through the generator during sampling. This is particularly costly when applied to modern DiT-based T2V models.

To add objects in specific locations in videos, most methods~\cite{ICLR2024_d9f8b5ab,Couairon_2023_ICCV,Chen_2026_CVPR,Bansal_2023_CVPR,10.1007/978-3-031-73113-6_3,Phung_2024_CVPR,ICLR2024_6f61f25a}, across both image and video generation domains, leverage the observation that these cross-attention layers are responsible for the objects' localization~\cite{Shen_2025_ICCV}. Thus, these strategies optimize for this similarity with gradients. However, we noted two primary drawbacks in computing gradients: (i) computing the mean cross-attention map explicitly is memory-intensive. In fact, modern neural network engines often use FlashAttention~\cite{dao2022flashattention} to avoid this explicit calculation; (ii) computing the gradient with respect to the input requires backpropagation through the entire network (billions of parameters) with very large spatio-temporal feature maps, necessitating VRAM-reducing techniques such as gradient checkpointing, multi-GPU sharding~\cite{10.14778/3611540.3611569}, or other engineering complexities, thereby significantly increasing inference time. For instance, in enormous architectures like Wan2.2~\cite{Wang2025WanOA}, computing the backward pass requires more than a single 80GB GPU. In contrast, users with consumer-grade GPUs, which have less than 32GB available VRAM, would not be able to utilize these methods. This motivates our setup, where we constrain T2V grounded generation to operate only in a training- and gradient-free environment. While existing gradient-free alternatives~\cite{Jain_2024_CVPR,NEURIPS2024_345208bd,Xu_2026_CVPR} avoid this overhead, they usually offer weaker or less reliable spatial control, especially in DiT architectures.

In this paper, we connect gradient-based methods with cross-attention manipulation techniques, proposing a training-free and gradient-free strategy that simulates the guiding behavior of their gradient-based counterparts. Our method, dubbed \underline{G}radient-free \underline{A}nalytical \underline{T}rajectory \underline{O}ptimization Video Generation, in short \METHODNAME, aims to optimize the global cross-attention maps between the target words and the spatio-temporal masks. To perform this guidance on-the-fly, we first analyze the standard cross-attention control loss and propose an efficient surrogate score. We then derive the analytical optimum of this objective, which provides explicit directions for increasing the alignment between target words and target regions. Finally, we introduce an injection mechanism that applies these directions directly within the transformer blocks, while respecting the geometry induced by query normalization.

Our contributions are as follows:
\begin{itemize}
\item We introduce \METHODNAME, a training-free and gradient-free framework for spatially grounded text-to-video generation that avoids backpropagation during inference in modern DiT-based video generators.
\item We reformulate cross-attention localization as an analytically solvable surrogate objective, deriving a closed-form solution that provides optimal directions without explicitly computing attention maps or gradients.
\item We propose a geometry-aware query injection mechanism that applies these analytical steering directions within transformer cross-attention modules while respecting the RMS normalization manifold, enabling stable and efficient inference-time control.
\end{itemize}

We demonstrate that \METHODNAME\ significantly improves quantitative localization metrics, whereas the tested baselines barely outperform the base model, highlighting the challenging nature of this task. To foster reproducibility of gradient-free and training-free approaches for grounded video generation, our code is available at \url{https://gato-vid.github.io/}.

\section{Related Work}

\subsection{Text-to-Video Generation}

Early Text-to-Video (T2V) generators~\cite{guo2023animatediff,10.1145/3680528.3687614,Chen_2024_CVPR} typically inflate existing text-to-image UNet models~\cite{10.1007/978-3-319-24574-4_28}, such as Stable Diffusion 1.5~\cite{Rombach_2022_CVPR}, and incorporate temporal layers to transfer feature information across the time dimension. While effective, these early models remained constrained along the temporal axis, which made them capable of generating videos of a few frames. More recently, with the introduction of DiTs~\cite{Peebles_2023_ICCV}, large-scale datasets~\cite{nan2025openvidm,wang2024internvid}, and massive training schemes~\cite{3692070.3692573}, DiT-based architectures have become the dominant paradigm. State-of-the-art models, including CogVideoX~\cite{yang2025cogvideox}, LTX-Video~\cite{HaCohen2026LTX2EJ}, HunyuanVideo 1.5~\cite{hunyuanvideo2025}, and Wan2.2~\cite{Wang2025WanOA}, leverage these architectures to generate high-quality, realistic videos from textual prompts. Nevertheless, their controllability remains restricted, as users are limited to guiding the generation using pure text.

Beyond text-only prompting, controllable video generation has also been studied through additional conditioning signals. A first line of work extends structural control to video, using frame-wise signals such as edge and depth maps or combining text, sketches, reference videos, and motion vectors to guide both spatial layout and temporal dynamics~\cite{chen2023controlavideo,wang2023videocomposer}. A second line focuses on motion-centric control, where users specify camera poses, camera movements, or object trajectories instead of relying solely on textual descriptions~\cite{wang2024motionctrl,he2024cameractrl,10.1145/3641519.3657481,geng2025motionprompting}. Rather than introducing new trainable layers, we achieve spatially grounded video generation by directly manipulating the T2V generator in a gradient-free manner.

\subsection{Spatially-Grounded Text-to-Image Control}

Instead of controlling the generation with mere text, users might want to generate content with objects in specific locations within a scene. In the text-to-image (T2I) domain, trainable approaches append modules on top of a frozen generator; for instance, GLIGEN~\cite{Li_2023_CVPR}, IFAdapters~\cite{Wu_2025_ICCV}, SpaText~\cite{Avrahami_2023_CVPR}, and BlobGEN~\cite{3692070.3693615} train new cross-attention layers to inject object descriptions into target locations. Conversely, training-free methods eliminate the need for parameter updates. Most methods rely on gradient-based optimization~\cite{Couairon_2023_ICCV,yang2024mftf,Xie_2023_ICCV,Chen_2024_WACV,Chen_2026_CVPR,Bansal_2023_CVPR,10.1007/978-3-031-73113-6_3,Phung_2024_CVPR,ICLR2024_6f61f25a}, maximizing the similarity between cross-attention maps and target object locations via backpropagation. This arises from the observation that these cross-attention layers are responsible for object localization~\cite{Shen_2025_ICCV}. While these methods do not require as much compute as video generators, backpropagating gradients through the network layers still incurs a notable computational overhead. Finally, the gradient-free literature is less extensive and typically can be categorized into two strategies: some methods~\cite{chen2024training,Ohanyan_2024_CVPR,10.1609/aaai.v38i6.28364} directly modify cross-attention layers to inject object segments or descriptions into their corresponding regions, while others~\cite{shirakawa2024noisecollage,pmlr-v202-bar-tal23a} utilize multi-inference pipelines to generate separate object estimates and subsequently merge them.

\subsection{Spatially-Grounded Text-to-Video Control}

Compared to the T2I domain, spatially grounded video generation remains less thoroughly explored. Mirroring text-to-image trends, existing approaches can be classified into the same three categories: training-based methods, gradient-based training-free strategies, and training- and gradient-free pipelines. Training-based strategies add new trainable modules on top of existing T2V models~\cite{Feng_2025_CVPR,dou2024gvdiffgroundedtexttovideogeneration}. Among gradient-based training-free methods, Lian \etal~\cite{ICLR2024_d9f8b5ab} optimize cross-attention maps between target words and input bounding boxes via backpropagation within the ModelScope~\cite{wang2023modelscopetexttovideotechnicalreport} framework. Within the gradient-free literature, attention manipulation is the dominant approach. For instance, Peekaboo~\cite{Jain_2024_CVPR} applies hard masking in both self- and cross-attention layers: in cross-attention, object text tokens attend strictly to their designated regions while the background attends elsewhere, whereas in self-attention, background and object video tokens are mutually isolated. Direct-a-Video~\cite{10.1145/3641519.3657481} follows a similar trend by constraining cross-attention maps within the ModelScope~\cite{wang2023modelscopetexttovideotechnicalreport} architecture. VideoTetris~\cite{NEURIPS2024_345208bd} operates akin to Regional Prompting~\cite{chen2024training} by inserting distinct object descriptions solely into their respective spatial regions. These previous methods have only been applied to 3D-UNet T2V models and have not been translated to modern DiT architectures. Closest to our work, SwitchCraft~\cite{Xu_2026_CVPR} tackles multi-event generation in Wan2.1~\cite{Wang2025WanOA} by projecting source queries into a target subspace to enhance similarity before reintegrating them. Unlike these methods, our approach is built on a theoretical foundation that optimizes a surrogate score on the fly in a completely gradient-free manner, while explicitly preserving the T2V model's geometric properties.
\section{Methodology}

In this section, we first define the problem and review standard gradient-based approaches. Next, we introduce our proposed surrogate objective and derive its analytical closed-form solution. Lastly, we explain how to incorporate this solution into each block of the DiT while maintaining its geometric properties.

\subsection{Preliminaries}\label{sec:formulation}

We begin by formally defining the main framework. Let $f_\theta$ be the T2V latent flow matching generator~\cite{Rombach_2022_CVPR,lipman2023flow}, and $\tau=\{e_1, \dots, e_{|\tau|},\, e_j\in \mathbb{R}^{d}\}$, be the input prompt tokenized and processed by the text encoder, consisting of $|\tau|$ vectors of dimension $d$, where $d$ is the hidden channel dimension of $f_\theta$. Let $x_t\in \mathbb{R}^{\mathrm{C}\times \mathrm{T}\times \mathrm{H} \times \mathrm{W}}$ be the input spatio-temporal video latent at time $t$, where $\mathrm{C}$, $\mathrm{T}$, $\mathrm{H}$, and $\mathrm{W}$ are the channel, time, height, and width dimensions, respectively. To synthesize a video using $f_\theta$, we iteratively refine $x_t$ according to the following equations
\begin{equation}\label{eq:gen}
\begin{split}
    v_t &= f_\theta(x_t,\tau,t) \\
    x_{t+\Delta t} &= \texttt{FlowSch}(x_t, v_t, t).
\end{split}
\end{equation}
Here, we initialize with Gaussian noise $x_0\sim\mathcal{N}(0,1)$, $\Delta t$ is the step size of the iterative process, and \texttt{FlowSch} is the flow matching scheduler. At the final timestep, \ie, $t=1$, the latent $x_1$ is decoded into pixel space using a pretrained variational autoencoder.

The training-free, gradient-free, spatially grounded T2V generation task aims to generate a video in which the object described in the prompt follows a given trajectory. Crucially, this control must be exerted during inference without altering the parameters of $f_\theta$ or computing gradients. To spatially ground the generative model, let $M$ denote the spatio-temporal indices tracking the target object's trajectory within the video volume. Finally, let $T$ be the set of textual indices pointing to the object in the tokenized prompt, $\tau$.

Diffusion transformers leverages native spatio-temporal transformer blocks~\cite{Peebles_2023_ICCV} to process the video data~\cite{yang2025cogvideox,Wang2025WanOA}.
Particularly, these architectures first concatenate text and video tokens and then process them through a sequence of blocks consisting of self-attention layers, followed by a feed-forward network. As noted by Shen~\etal~\cite{Shen_2025_ICCV}, a sub-operation within self-attention is the cross-attention function, where text data is injected into video tokens. Thus, regardless of the architecture, whether it is a UNet~\cite{guo2023animatediff} or a DiT~\cite{Peebles_2023_ICCV}, the fundamental component for grounding visual features with textual information is the cross-attention mechanism:
\begin{equation}
\begin{split}
    A(Q_i,K_i) &= \texttt{softmax}\left(\frac{Q_iK_i^\top}{\sqrt{d}}\right)\\
    \texttt{cross-attn}&(Q_i,K_i,V_i) = A(Q_i,K_i)V_i,
\end{split}
\end{equation}
where $Q_i\in \mathbb{R}^{\mathrm{T}\mathrm{H}\mathrm{W}\times d}$, $K_i\in \mathbb{R}^{|\tau|\times d}$, and $V_i\in \mathbb{R}^{|\tau|\times d}$ represent the query (video), key (text), and value (text) sequences, respectively, for the $i$-th block of the network. In modern DiT architectures, to generate the queries $Q_i$, the sequence features are mapped via a linear projection layer. Then, the RMS Normalization~\cite{NEURIPS2019_1e8a1942} (\cref{eq:rmsnorm}) projects all individual vectors $q\in Q_i$, $q\in \mathbb{R}^d$, onto a hyper-ellipsoid surface:
\begin{equation}\label{eq:rmsnorm}
    q \leftarrow \sqrt{d}\frac{q\odot\gamma}{\|q\|},
\end{equation}
where $\odot$ denotes the element-wise multiplication and $\gamma\in \mathbb{R}^d$ is a learnable vector.

Gradient-based algorithms generate an object in a location $M$ by first passing $x_t$ and $\tau$ through $f_\theta$ to extract the vision queries and text keys:
\begin{equation}
    (v_t, \{Q_i,K_i\}_{i=1}^B) = f_\theta(x_t, \tau,t),
\end{equation}
where $v_t$ is the model output, and $B$ is the total number of blocks in $f_\theta$. Second, they compute the mean attention map over the $B$ blocks:
\begin{equation}\label{eq:mean-attn-map}
    \bar{A} = \frac{1}{B} \sum_{i=1}^B A(Q_i, K_i).
\end{equation}
Third, they compute the gradients of a loss function, $\ell$, between the mean attention $\bar{A}$, the target words, $T$, and the target location $M$ with respect to the input $x_t$. This function $\ell$ is typically a Dice or $\ell_2$ loss between $\bar{A}$, which is averaged over each target text token, and $M$. Then, the latent $x_t$ is updated with: 
\begin{equation}\label{eq:backprop}
    x_t' = x_t - \mu \nabla_{x_t} \ell(\bar{A},M,T),
\end{equation}
where $\mu$ is the learning rate. Finally, the subsequent denoising step is computed using $x_t'$ instead of $x_t$, or it is refined iteratively with more gradient steps. In this paper, our goal is to derive an efficient surrogate score that analytically captures an optimal direction for decreasing $\ell$ instead of optimizing this loss directly with gradients.

\subsection{Building a Simpler Score}

\begin{figure}[t]
    \centering
    \begin{subfigure}[b]{0.49\linewidth}
        \includegraphics[width=\linewidth]{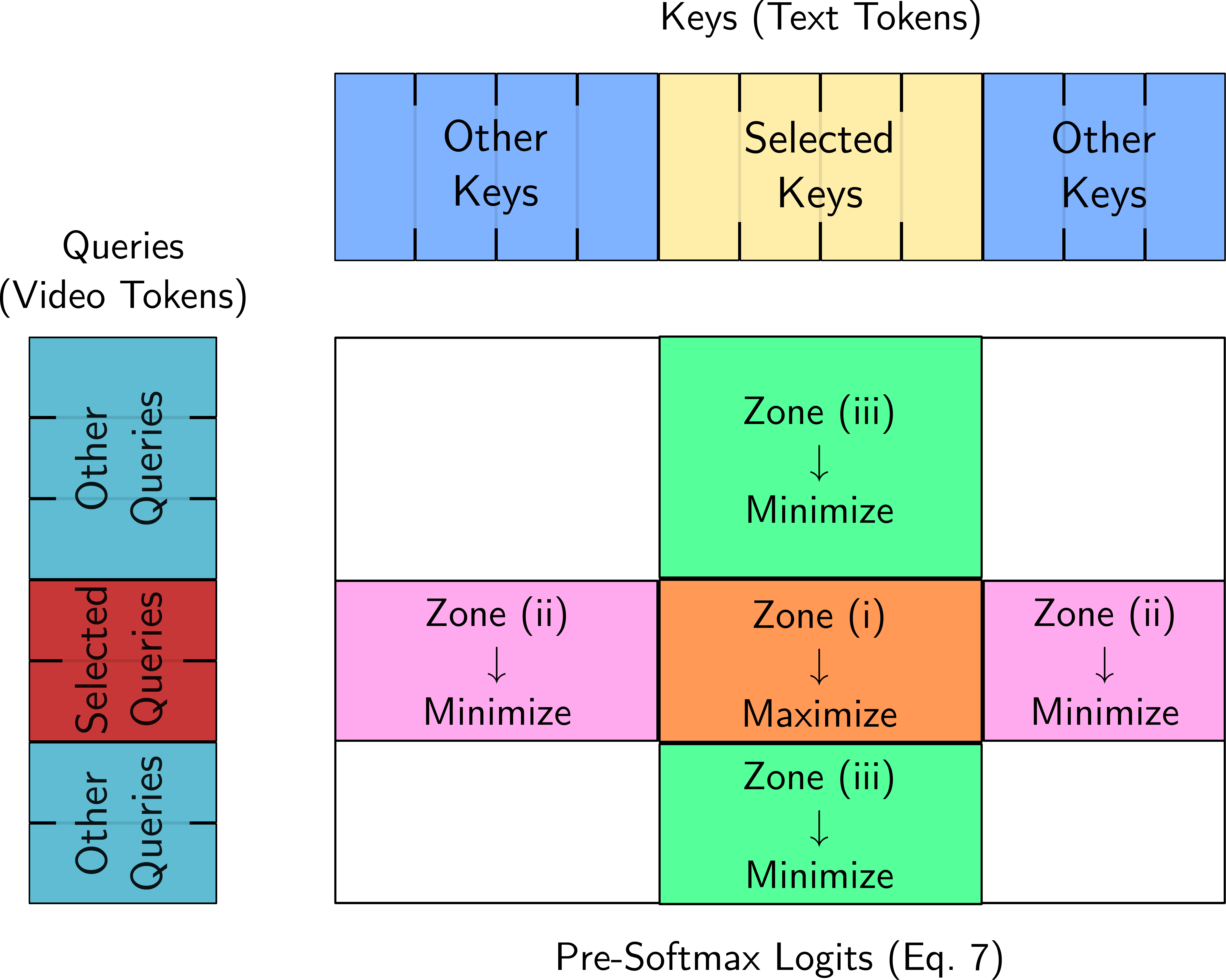}
        \caption{}
        \label{fig:toyexamplea}
    \end{subfigure}
    \begin{subfigure}[b]{0.49\linewidth}
        \includegraphics[width=\linewidth]{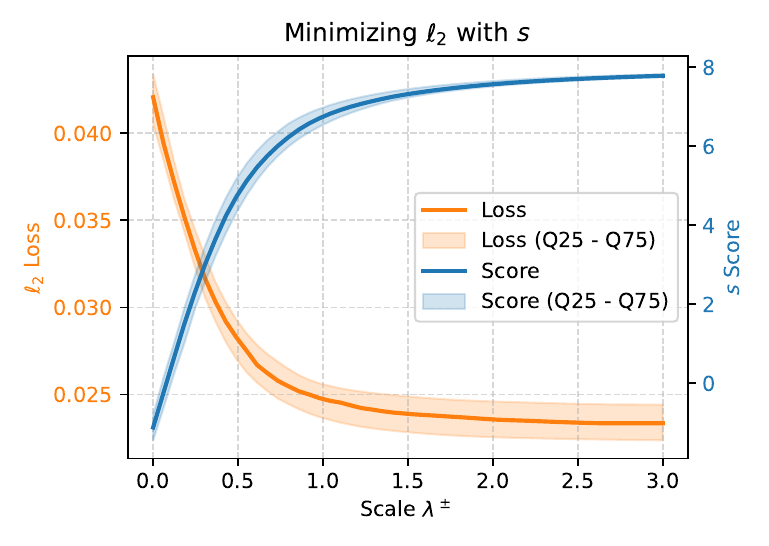}
        \caption{}
        \label{fig:toyexampleb}
    \end{subfigure}
    \caption{\textbf{(a) Target Regions:} Cross-attention scores targeted with our score on \cref{eq:score-non-sep}. $s$ searches to: {\color{BurntOrange}(i)} increase the logits of the target spatio-temporal regions and selected text tokens, {\color{Rhodamine}(ii)} decrease the logits of the target region and complementary text tokens, and {\color{OliveGreen}(iii)} reduce the logits of the regions outside the mask and the chosen text tokens. \textbf{(b) Loss Minimization:} Toy example exemplifying that our on-the-fly score optimization minimizes an $\ell_2$ loss. The plot illustrates how increasing the $\lambda^\pm$ parameter gradually decreases the $\ell_2$ error over different runs. The ``Q25--Q75'' shaded area represents the 25th--75th quantile range.}
    \label{fig:toyexample}
    \vspace{-3mm}
\end{figure}

Our first contribution is an analytically solvable alternative score function to approximate the behavior of the loss function. Essential to the cross-attention mechanism is the softmax operation, a highly non-linear function that normalizes the cross-attention scores. To derive an exact analytical solution for the alternative loss function, we need a proxy for the softmax to circumvent the non-linearity. Formally, if $q_j \in Q_i$ and $k_l \in K_i$ are a specific query token and a key token, respectively, then, the cross-attention between $q_j$ and $k_l$ is:
\begin{equation}\label{eq:softmax}
    A(q_j, K_i) = \left[ \dots, \frac{\texttt{exp}( \nicefrac{q_j \cdot k_l}{\sqrt{d}})}{\sum_{k \in K_i} \texttt{exp}(\nicefrac{q_j \cdot k}{\sqrt{d}})}, \dots \right].
\end{equation}
To maximize the attention between the $l$-th text token ($k_l$) and the $j$-th query ($q_j$), we can either maximize $q_j \cdot k_l$ or minimize $\sum_{k \in K, k \neq k_l} \texttt{exp} (\nicefrac{q_j \cdot k}{\sqrt{d}})$. Fortunately, to minimize this second term, we can directly work with each $q_j \cdot k_l$ individually as $\texttt{exp} (\nicefrac{q_j \cdot k}{\sqrt{d}})$ is strictly increasing with respect to $q_j \cdot k_l$. Thus, we can directly utilize the pre-softmax operation (\ie, the logits),
\begin{equation}\label{eq:core}
    A'(Q,K) = QK^\top,
\end{equation}
and use it as a proxy for optimizing the attention. Consequently, we build our score function directly on top of \cref{eq:core}.
To increase attention in $M$, we \textit{maximize} the following score:
\begin{equation}\label{eq:score-non-sep}
    s = \underbrace{\frac{\sum_{j \in M} \sum_{l \in T} q_j \cdot k_l}{|M| |T|}}_{\color{BurntOrange} (i)}   
    - \underbrace{\frac{\sum_{j \in M} \sum_{l \in T^c} q_j \cdot k_l}{|M| |T^c|}}_{\color{Rhodamine}(ii)}
    - \underbrace{\frac{\sum_{j \in M^c} \sum_{l \in T} q_j \cdot k_l}{|M^c| |T|}}_{\color{OliveGreen}(iii)},
\end{equation}
where $M^c$ and $T^c$ are the complement sets of indices of $M$ and $T$, respectively. This score consists of three terms (from left to right): {\color{BurntOrange}(i)} a factor to increase the inner product, and thus the attention, between target regions and target words, {\color{Rhodamine}(ii)} a component reducing the influence of negative keys in target regions, and {\color{OliveGreen}(iii)} an element reducing the influence of positive keys outside target regions. In \cref{fig:toyexamplea} we visually represent all regions {\color{BurntOrange}(i)}, {\color{Rhodamine}(ii)}, and {\color{OliveGreen}(iii)}. Analogous to the original loss function $\ell$, maximizing $s$ implicitly minimizes terms {\color{Rhodamine}(ii)} and {\color{OliveGreen}(iii)} while maximizing term {\color{BurntOrange}(i)}, but operates completely in the logit space.

Since the double sum is separable, we can factorize \cref{eq:score-non-sep} into:
\begin{equation}\label{eq:score}
    s = \mathcal{Q}(M) \cdot (\mathcal{K}(T)
    - \mathcal{K}(T^c))
    - \mathcal{Q}(M^c) \cdot \mathcal{K}(T),
\end{equation}
where $\mathcal{Q}(M) \in \mathbb{R}^d$ and $\mathcal{K}(T) \in \mathbb{R}^d$ are defined as:
\begin{equation}
    \mathcal{Q}(M)= \frac{\sum_{j \in M} q_j}{|M|} , \quad \mathcal{K}(T)=\frac{\sum_{l \in T} k_l }{|T|}.
\end{equation}
This factorization avoids the explicit attention map calculation, reducing the complexity from $\mathcal{O}(|Q| \cdot |K|)$ to three vector dot products. In \cref{fig:toyexampleb}, we provide a toy example illustrating how maximizing $s$ acts as a proxy for minimizing a standard $\ell_2$ localization loss - see \cref{sec:on-the-fly} for more details.

\subsection{On-the-fly Score Optimization}\label{sec:on-the-fly}

To avoid any gradient-based optimization, we first analytically maximize our proposed score in \cref{eq:score} by finding the vectors $\mathcal{Q}(M)$ and $\mathcal{Q}(M^c)$ that maximize $s$. Suppose $b^+$ and $b^-$ are unit vectors replacing $\mathcal{Q}(M)$ and $\mathcal{Q}(M^c)$, respectively. The score $s$, which is now a function of $(b^+,b^-)$, becomes:
\begin{equation}
    s(b^+,b^-) = b^+ \cdot \left(\mathcal{K}(T) - \mathcal{K}(T^c)\right) - b^- \cdot \mathcal{K}(T).
\end{equation}
Consequently, $s$ is maximized when $b^+$ is collinear with $(\mathcal{K}(T) - \mathcal{K}(T^c))$ and $b^-$ is parallel to $(-\mathcal{K}(T))$. Thus, we can maximize the attention map by simply replacing these vectors (now referred to as biases) with the query sequence. Accordingly, we set $b^+$ and $b^-$ as
\begin{equation}\label{eq:biases}
    b^+ = \frac{\mathcal{K}(T) - \mathcal{K}(T^c)}{\|\mathcal{K}(T) - \mathcal{K}(T^c)\|} \quad \text{and} \quad b^-=-\frac{\mathcal{K}(T)}{\|\mathcal{K}(T)\|}.
\end{equation}

Now, for each cross-attention block, we have an optimal direction vector. Yet, directly replacing the queries $q_j$ with the optimal biases ($q_j \leftarrow b^\pm$) or naively adding the biases ($q_j \leftarrow b^\pm + q_j$) disrupts the network's internal representations due to magnitude mismatches. Here, $b^\pm$ is either $b^+$ or $b^-$. As the query vectors lie on the RMSNorm ellipsoid (\cref{eq:rmsnorm}), we need to take into account this geometric property to remain closer to the distribution. Thus, we modify the queries as follows:
\begin{equation}\label{eq:inj-proj}
    q_j \leftarrow \begin{cases} 
    \frac{\sqrt{d}}{\|(\lambda^+ b^+ \|q_j\| + q_j) \oslash \gamma\|}  (\lambda^+ b^+ \|q_j\| + q_j) & \text{if } j \in M \\
    \frac{\sqrt{d}}{\|(\lambda^- b^- \|q_j\| + q_j) \oslash \gamma\|}  (\lambda^- b^- \|q_j\| + q_j) & \text{if } j \in M^c 
    \end{cases},
\end{equation}
where $\oslash$ is the element-wise division. This renormalization consists of three main components: (i) We multiply $b^\pm$ by $\|q_j\|$ to make the impact relative to $q_j$. 
(ii) We scale the bias with a hyperparameter $\lambda^\pm$ to control the impact of the bias $b^\pm$. Here, $\lambda^\pm$ plays a similar role to the learning rate $\mu$ in gradient-based approaches (\cref{eq:backprop}). 
(iii) The constant $\frac{\sqrt{d}}{\|(\lambda^\pm b^\pm\|q_j\| + q_j) \oslash \gamma\|}$ projects the vector back to the ellipsoid surface of the RMS Normalization in \cref{eq:rmsnorm}.

For the background tokens ($j \in M^c$), adding the negative bias $b^-$ can distort the existing background features if the bias aligns too closely with the original query $q_j$. To prevent this distortion, we isolate the component of the bias that is strictly orthogonal to the query, ensuring the injection acts purely as a directional adjustment without overwriting the original feature content.

Formally, we define the orthogonal projection of a vector $a$ onto the orthogonal complement of $b$ as:
\begin{equation}
    P(a,b) = a - \left(a \cdot \frac{b}{\|b\|}\right) \frac{b}{\|b\|}.
\end{equation}
Using this projection, our final formulation for the query modification becomes:
\begin{equation}
    q_j \leftarrow \begin{cases}
    \frac{\sqrt{d}}{\left\| (\lambda^+ b^+ \|q_j\| + q_j) \oslash \gamma \right\|} \left(\lambda^+ b^+ \|q_j\| + q_j\right) & \text{if } j \in M, \\
    \frac{\sqrt{d}}{\left\| \left(\lambda^- P(b^- \|q_j\|, \, q_j) + q_j\right) \oslash \gamma \right\|} \left(\lambda^- P(b^- \|q_j\|,  q_j) + q_j\right) & \text{if } j \in M^c
    \end{cases}.
\end{equation}

As a final adjustment, we note that the cross-attention activations typically peak at the semantic center of an object and attenuate toward its boundaries. While we correctly insert the positive bias into each query, uniformly applying it yields an unnaturally homogeneous attention distribution across the target region. To circumvent this problem, we modulate the guidance scale spatially by fitting a normalized 2D Gaussian distribution to the bounding box of each frame. Subsequently, we multiply it by $\lambda^+$. Hence, $\lambda^+$ varies spatially now than remaining fixed for every query inside the mask.

To validate empirically that maximizing our gradient-free surrogate score $s$ with our injection mechanism in \cref{eq:inj-proj} effectively acts as a proxy for minimizing a standard $\ell_2$ localization loss, we conduct a controlled toy experiment. Specifically, we instantiate a three-layer cross-attention model with randomly initialized weights and apply our directional query steering mechanism across a range of guidance intensities $\lambda^\pm$. We conduct this process 100 times, storing the $(\lambda^\pm, \ell_2, s)$ values and plotting both $(\lambda^\pm, s)$ and $(\lambda^\pm, \ell_2)$ sequences. As illustrated in \cref{fig:toyexampleb}, increasing $\lambda^\pm$ monotonically reduces the $\ell_2$ loss between the resulting attention maps and the given target spatial layout $M$. Our analytical approximation yields highly localized attention maps, offering an efficient trade-off by bypassing the heavy  backpropagation during inference.

\begin{algorithm}[t]
\caption{\textbf{Bias Injection:} First, the positive and negative directions are computed using the selected text tokens. Then, the biases are injected back into the queries, which are then reprojected onto the RMSNorm ellipsoid. Finally, the newly modified tokens through the cross-attention.}
\label{alg:forward}
\begin{algorithmic}
\Require Generator $f_\theta$, $x$ video features, $t$ current step, prompt features $\tau\in\mathbb{R}^{|\tau|\times n}$, target map $M$, selected words $T$, $S_B$ set of blocks where we inject the bias.
\State $x \leftarrow f\texttt{.tokenize}(x) \in\mathbb{R}^{\mathrm{THW}\times n}$
\For{$b$ in $f_\theta$\texttt{.blocks}}
    \State $x\leftarrow x + b.\texttt{Self-Attention}(x)$
    \State $q, k, v \leftarrow b\texttt{.norm\_q}(b\texttt{.to\_q}(x)), b\texttt{.norm\_k}(b\texttt{.to\_k}(\tau)), b\texttt{.to\_v}(\tau)$
    \If{$b \in S_B$}
        \State $b^+ = \texttt{Normalize}\big(\texttt{mean}(k[T]) - \texttt{mean}(k[T^c])\big)$
        \State $q' \leftarrow q[M]$ \Comment{Get queries inside the mask}
        \State $q' \leftarrow q' + \lambda^+ \|q'\| b^+$  \Comment{Add positive bias}
        \State $q' \leftarrow \frac{\sqrt{n}}{\|q'\oslash b\texttt{.norm\_q}.\gamma\|} q'$  \Comment{Normalization to hyper-ellipsoid}
        \State $q[M] \leftarrow q'$
        \State $b^- = \texttt{Normalize}(\texttt{mean}(k[T^c]))$
        \State $q' \leftarrow q[M^c]$ \Comment{Get queries outside the mask}
        \State $q' \leftarrow q' - \lambda^+ \|q'\| b^-$  \Comment{Add negative bias}
        \State $q' \leftarrow \frac{\sqrt{n}}{\|q'\oslash b\texttt{.norm\_q}.\gamma\|} q'$  \Comment{Normalization to hyper-ellipsoid}
        \State $q[M^c] \leftarrow q'$
    \EndIf
    \State $x \leftarrow x + b.\texttt{cross-attention}(q,k,v)$
    \State $x\leftarrow x + b.\texttt{FFN}(x)$
    \State \Return $f.\texttt{detokenize}(x)$
\EndFor
\end{algorithmic}
\end{algorithm}

In a nutshell, to generate a spatially grounded video with \METHODNAME{}, we first sample the noisy latent variables with Gaussian noise $x_0$. Next, for the first few iterations and the first blocks of the generative model, we extract the target biases $(b^+, b^-)$ and inject them to the queries following \cref{eq:inj-proj}, summarized in Alg.~\ref{alg:forward}. For the remaining iterations, we adhere to the original iterative process in \cref{eq:gen}.

\section{Experimentation}

\subsection{Implementation Details}

We implemented \METHODNAME\ on Wan2.2~\cite{Wang2025WanOA}, the best open-sourced T2V generative model. We used our injection mechanism in the first 20 blocks of the transformer architecture during the first $15\%$ of the iterations. Furthermore, we set $\lambda^\pm=1.5$ with a linear decay. As our baseline choices, we selected the most representative methods available in the literature for gradient-free and training-free grounded T2V generation. These are Peekaboo~\cite{Jain_2024_CVPR}, VideoTetris~\cite{NEURIPS2024_345208bd}, and SwitchCraft~\cite{Xu_2026_CVPR}. To mitigate the effect of using other architectures, we adapted them all to Wan2.2 for a fair comparison. Moreover, we performed a hyperparameter search in all baselines to adjust for the different backbone. Additionally, we evaluated Wan2.2 without grounding to establish a quality reference. All generated videos have 81 frames with a spatial resolution of 480$\times$832 pixels using 30 steps flow matching steps. Finally, all experiments were performed on a single NVIDIA H100 (80GB) GPU.

\subsection{Evaluation Setup}

\paragraph{Dataset} To benchmark \METHODNAME\ against concurrent approaches, we constructed two dedicated evaluation datasets. On the one hand, we leveraged Gemini~\cite{geminiteam2025geminifamilyhighlycapable} to generate 25 diverse prompts covering a variety of objects. For each prompt, we generated four ground-truth videos using Wan2.2 across different random seeds and visually verified the results. We then applied SAM~3~\cite{carion2026sam} to extract the corresponding reference bounding boxes for the target objects. During evaluation, we executed each prompt and bounding box configuration using four independent random seeds. In total, our benchmark comprises 400 generated videos. We name this set \textit{Set 1}. On the other hand, we leveraged ChatGPT~\cite{openai2024gpt4technicalreport} to generate 100 distinct prompts, creating synthetic bounding boxes of various sizes, shapes, and movement patterns (linear, spiral, circular, Z/N-shaped movements) for each prompt. This set is inherently more challenging yet closer to our application, as the target locations conflict with the model's typical movement patterns. Consistent with \textit{Set 1}, we ran each prompt across four different seeds, yielding a total of 400 videos. We refer to this collection as \textit{Set 2}.

\paragraph{Evaluation Metrics} We evaluate our approach along two dimensions: spatial alignment and overall video quality. Spatial alignment metrics assess whether the target object was accurately generated within the specified bounding box while respecting its dimensions. To achieve this, we follow an evaluation protocol similar to that in~\cite{Jain_2024_CVPR}. Given a set of trajectory boxes, the textual description of the object, and the generated video, we extract object masks using SAM~3~\cite{carion2026sam} prompted with the target object's prompt. Then, we compute two metrics to measure the similarity with the input bounding boxes: (i) the Intersection over Union (IoU) between the bounding box of the generated object and the target box (higher is better) and (ii) the Euclidean distance between the center of mass of the generated object and the center of the target bounding box, divided by the maximum possible distance, named Center Distance (CD); lower is better. Both CD and IoU are computed exclusively for videos with successful SAM~3 object detections. To complement this, we introduce a Success Rate (SR) metric to track the proportion of generated videos that contain the target object according to SAM~3. This metric is especially useful for identifying methods that fail to generate the target object altogether.

To evaluate overall video generation quality (\eg, temporal consistency, visual quality), we adopt VBench~\cite{huang2023vbench} metric suite. Specifically, we evaluate five core dimensions: Subject Consistency (SC), Background Consistency (BC), Dynamic Degree (DD), Aesthetic Quality (AQ), and Imaging Quality (IQ). SC and BC measure the temporal stability of objects and backgrounds across video frames, ensuring they do not drastically shift or warp. DD quantifies the amount of motion and dynamic variation within the video. Finally, AQ assesses the stylistic and compositional appeal of each frame, while IQ penalizes visual artifacts such as noise, blur, or distortions. For all metrics, higher values indicate better performance. Ideally, each method yields scores comparable to the vanilla model; a substantial change signals a noticeable shift from the base behavior.

\paragraph{Quantitative Result}

\begin{table}[t]
    \centering
    \scriptsize
    \begin{tabular}{C{0.4cm}|L{2.4cm}|C{1.0cm}C{1.0cm}C{1.0cm}|C{1.0cm}C{1.0cm}C{1.0cm}C{1.0cm}C{1.0cm}} \toprule
        &\multirow{2}{*}{\textbf{Method}} & \multicolumn{3}{c|}{\textbf{Localization}} & \multicolumn{5}{c}{\textbf{VBench}}\\
                    & & CD ($\downarrow$)     & IoU ($\uparrow$) & SR ($\uparrow$)  & SC ($\uparrow$)  & BC  ($\uparrow$)   & DD ($\uparrow$)  & AQ ($\uparrow$) & IQ ($\uparrow$)  \\ \midrule
        \multirow{5}{*}{\rotatebox[origin=r]{90}{\textit{Set 1}}}
        &Wan2.2 T2V  & 0.138 & 0.154 & 89.3\% & 0.974& 0.971  & 0.248&\textbf{0.631}&0.666 \\ \cmidrule{2-10}
        &VideoTetris$^*$& 0.132 & 0.160 & 85.7\% & 0.973 & \textbf{0.972} & 0.260 & 0.630      & 0.673 \\
        &SwitchCraft & 0.138 & 0.152 & 88.7\% & 0.973 & \textbf{0.972} & 0.248 & 0.629      & 0.664 \\
        &Peekaboo$^*$   & 0.103 & 0.249 & 85.5\% & 0.970 & 0.966 & \textbf{0.310} & 0.597      & \textbf{0.677} \\ \cmidrule{2-10}
&\METHODNAME\ (Ours) & \textbf{0.059} & \textbf{0.363} & \textbf{89.7\%} & \textbf{0.975} & 0.964 & 0.113 & 0.551      & 0.670\\ \midrule
\multirow{5}{*}{\rotatebox[origin=r]{90}{\textit{Set 2}}}
        &Wan2.2 T2V  & 0.198 & 0.124 & 74.2\% & 0.957 & 0.960 & 0.395 & \textbf{0.613} & 0.682 \\ \cmidrule{2-10}
        &VideoTetris$^*$& 0.195 & 0.136 & 69.7\% & \textbf{0.960} & \textbf{0.962} & 0.402 & 0.612 & \textbf{0.698} \\
        &SwitchCraft & 0.196 & 0.128 & \textbf{75.5\%} & 0.956 & 0.960 & 0.393 & 0.611 & 0.678 \\
        &Peekaboo$^*$   & 0.193 & 0.171 & 49.2\% & 0.929 & 0.937 & \textbf{0.750} & 0.510 & 0.679 \\ \cmidrule{2-10}
&\METHODNAME\ (Ours) & \textbf{0.121} & \textbf{0.324} & 74.7\% & 0.952 & 0.951 & 0.360 & 0.557 & 0.690 \\ \bottomrule
    \end{tabular}
    \caption{\textbf{Quantitative Localization and Quality Metrics:} \METHODNAME\ outperforms every baseline on the localization metrics with a trade-off. Its superior alignment comes at the cost of generation quality, as shown by the VBench metrics. All methods marked with $^*$ were re-implemented on Wan2.2.}
    \label{tab:sota}
    \vspace{-3mm}
\end{table}

We report the quantitative comparison between \METHODNAME\ and various baselines in \cref{tab:sota}. Importantly, no baseline demonstrates robust localization capabilities. Peekaboo leads with a modest IoU on both datasets, while all other configurations perform comparably to the standard T2V model. In contrast, \METHODNAME\ outperforms all existing baselines by a large margin across all spatial alignment metrics, while being on par in the SR metric. Several key findings from these results merit closer inspection. First, we observe that directly adapting training-free, gradient-free strategies~\cite{NEURIPS2024_345208bd,Jain_2024_CVPR} originally designed for 3D-UNet architectures to modern DiT models yields poor results. Specifically, VideoTetris~\cite{NEURIPS2024_345208bd} performs comparably to the vanilla T2V model, completely failing to ground the target objects in their specified locations. Conversely, Peekaboo~\cite{Jain_2024_CVPR} achieves the highest alignment scores among the baselines; however, it introduces visual artifacts, which we analyze qualitatively in \cref{sec:qualitative}. Second, SwitchCraft~\cite{Xu_2026_CVPR} - originally built for Wan2.1 - exhibits a similar limitation to VideoTetris, matching its performance to the unconstrained Wan2.2 baseline. We attribute this to SwitchCraft's inability to suppress the influence of target text tokens outside the spatial mask; our ablation study (\cref{sec:ablations}) shows that a negative component is necessary to improve alignment metrics. Third, the VBench evaluation indicates that all methods perform similarly across most dimensions.
However, a decline in \METHODNAME{}'s DD and AQ scores suggests that our spatial constraints can lead to more static compositions. This stems from the background degradation and reduced scene dynamics of certain videos. Peekaboo exhibits a similar trade-off, albeit less severely. The difference in the DD metric to the T2V vanilla model suggests that Peekaboo creates excessive movement, and the decreased AQ indicates lower-quality videos. This final discussion pinpoints that better localization comes at the cost of quality. Finally, we underscore the inherent difficulty of precise training-free gradient-free spatial control.


Next, we measure the computational efficiency of \METHODNAME{} by recording the average inference time across 50 runs. \METHODNAME{} introduces minimal runtime overhead of 0.4\% compared to Vanilla Wan2.2. By comparison, VideoTetris adds 6.20\%, SwitchCraft adds 32.80\%, and Peekaboo adds 92.13\%. However, since these methods (including ours) are only applied during the initial sampling steps, their impact on the total generation time is modest to negligible. In contrast, performing a single backpropagation step on one of Wan2.2's two transformers increases inference time by 300.31\% and requires an additional 26 GB of VRAM (totaling 59 GB) with only one transformer loaded in memory and without computing an attention map loss. When computing the explicit attention maps and the attention loss on an 80GB GPU, we received out-of-memory errors. Thus, these methods require CPU offloading strategies, multiple GPUs, and gradient checkpointing. This makes gradient-based approaches unattractive for users with limited compute budgets.

\subsection{Qualitative Results}\label{sec:qualitative}

\begin{figure}[t]
    \centering
    \includegraphics[width=0.95\linewidth]{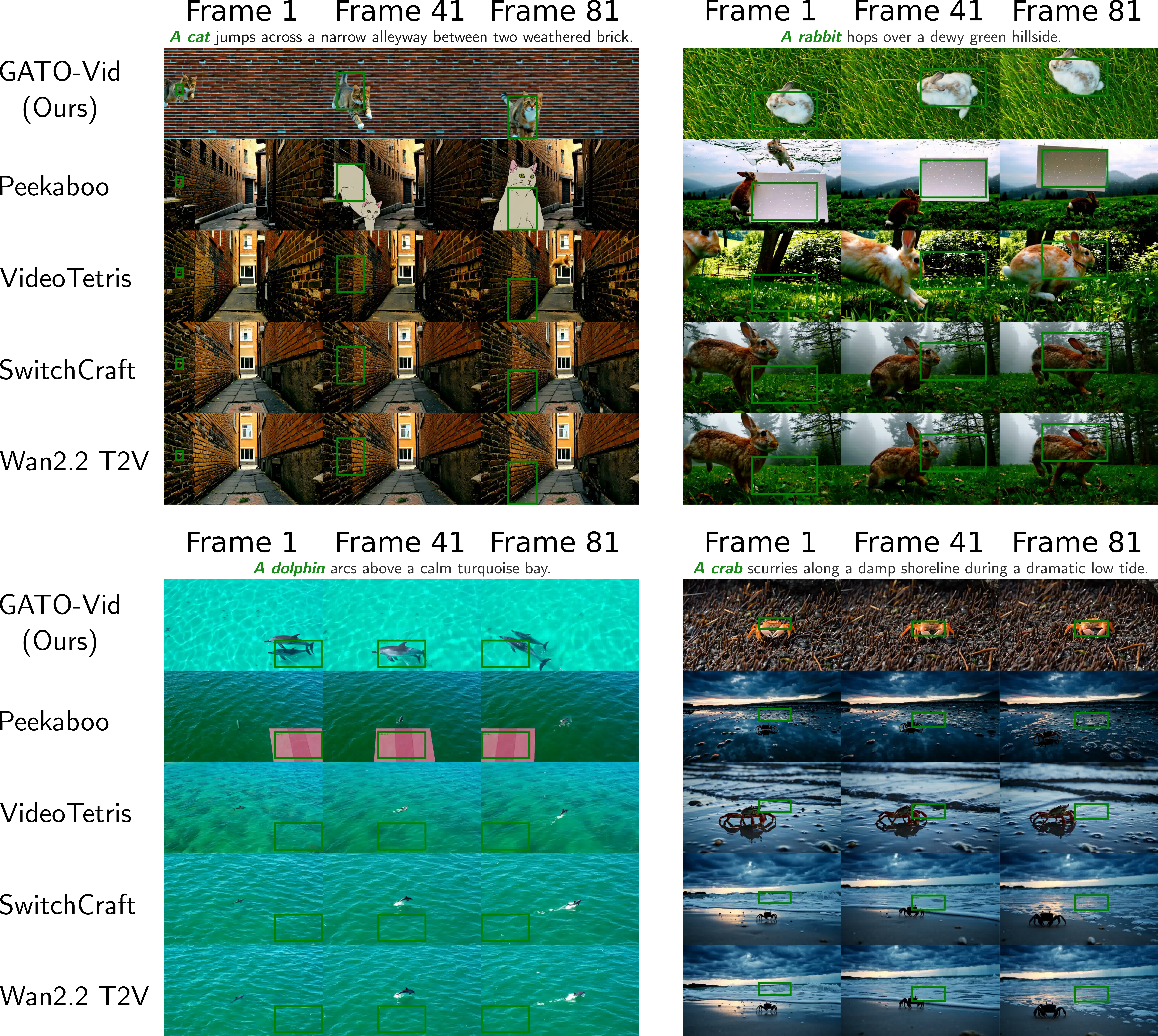}
    \caption{\textbf{Qualitative Comparison:} From top to bottom, rows display results for \METHODNAME, Peekaboo, VideoTetris, SwitchCraft, and the vanilla T2V model, all sharing identical random seeds. Prompts are provided on top of each video, with the target object and bounding boxes highlighted in green.}
    \label{fig:qualitative}
    \vspace{-3mm}
\end{figure}

To complement our study, we provide visual examples in \cref{fig:qualitative}. First, Vanilla Wan2.2, VideoTetris, and SwitchCraft produce similar results. They preserve the original quality because these algorithms are not strong enough to steer the generation toward the goal. Next, Peekaboo results are of good quality; however, the hard masking introduces visual artifacts in certain regions of the outputs. 
In contrast, our method succeeds where the baselines fail by successfully generating and localizing the target object within the specified region at the expense of background quality.
For more visualization, we refer to our \href{https://gato-vid.github.io/}{webpage}.

\subsection{Ablation Studies}\label{sec:ablations}


\begin{table}[t]
    \centering
    \scriptsize
    \begin{tabular}{L{3cm}|C{1.0cm}C{1.0cm}C{1.0cm}|C{1.0cm}C{1.0cm}C{1.0cm}C{1.0cm}C{1.0cm}} \toprule
        \multirow{2}{*}{\textbf{Ablation}} & \multicolumn{3}{c|}{\textbf{Localization}} & \multicolumn{5}{c}{\textbf{VBench}} \\
                         & CD ($\downarrow$)     & IoU ($\uparrow$)  & SR ($\uparrow$)  & SC ($\uparrow$)  & BC  ($\uparrow$)   & DD ($\uparrow$)  & AQ ($\uparrow$) & IQ ($\uparrow$) \\ \midrule
    Wan2.2 T2V           & 0.138 & 0.154 & 89.3\% & 0.974& \textbf{0.971}  & 0.248&\textbf{0.631}&0.666 \\ \midrule
    (i) No Projection    & \textbf{0.059} & 0.355 & 87.5\% &0.965& 0.957  & 0.090&0.541&0.646\\
(ii) No Gaussian Fitting & \textbf{0.059} & \textbf{0.375} & 72.0\% &0.969& 0.960  & 0.093&0.535&0.662\\
    (iii) No Foreground  & 0.104 & 0.252 & \textbf{89.7\%} &\textbf{0.977}& 0.968  & 0.158&0.558&0.660\\
    (iv) No Background   & 0.105 & 0.241 & \textbf{89.7\%} &0.972& 0.970  & \textbf{0.258}&0.628&\textbf{0.679}\\ \midrule
\METHODNAME\ (Ours) & \textbf{0.059} & 0.363 & \textbf{89.7\%} & 0.975 & 0.964 & 0.113 & 0.551      & 0.670\\ \bottomrule
    \end{tabular}
    \caption{\textbf{Ablation Study:} Effect of systematically removing each proposed component to validate its advantages. Eliminating any part of \METHODNAME\ degrades localization accuracy, which empirically validates all of our proposed components.}
    \label{tab:ablations}
    \vspace{-3mm}
\end{table}
\begin{figure}[t]
    \centering
    \includegraphics[width=.95\linewidth]{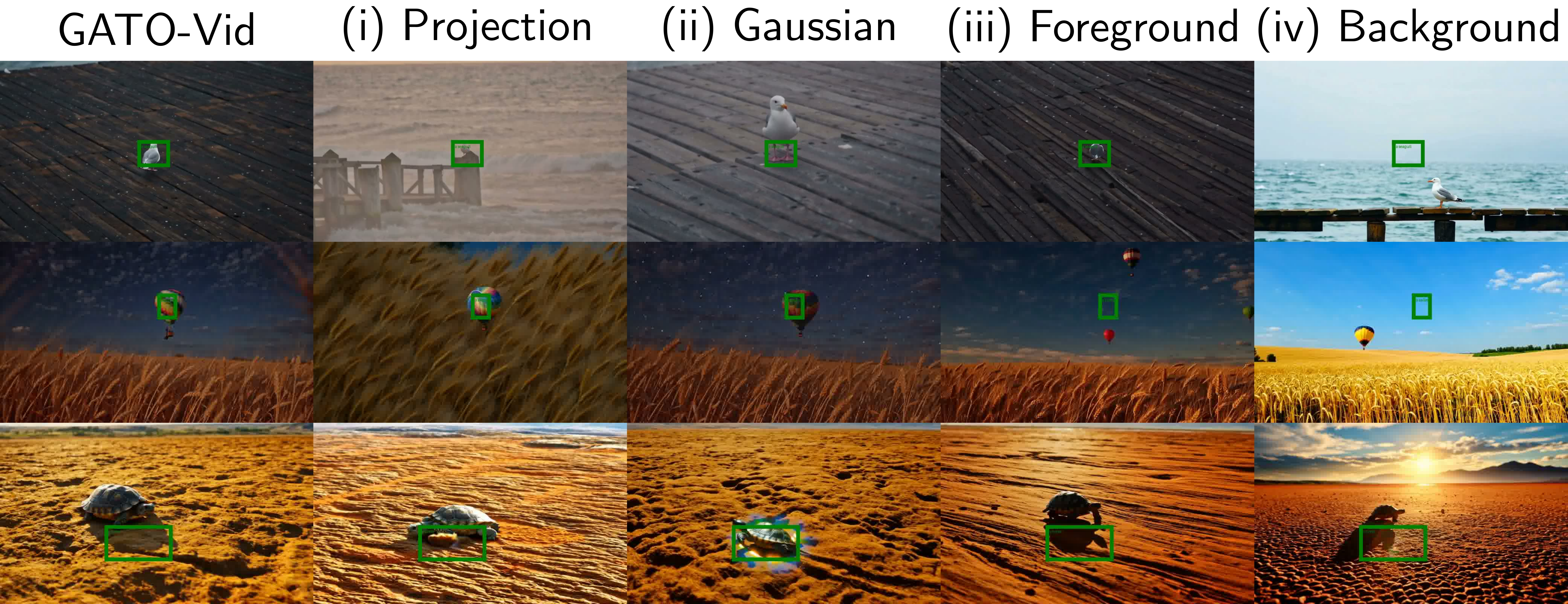}
    \caption{\textbf{Ablation Visualization:} Representative frames from three videos illustrating the visual degradation under different ablation settings. Removing either the reprojection mechanism (i) or the Gaussian filtering (ii) introduces noticeable visual artifacts. Conversely, omitting the positive (iii) or negative biases (iv) leads to localization errors.}
    \label{fig:ablation}
    \vspace{-3mm}
\end{figure}

We conducted an ablation study to demonstrate the advantages of each proposed component. To do so, we performed the following four ablations: (i) We removed the projection to the ellipsoid, found in \cref{eq:inj-proj}. Thus, we replaced the injection with a naive sum ($q_j \leftarrow q_j + \lambda^\pm \|q_j\| b^\pm$). (ii) We removed the Gaussian filtering, used to mimic the smooth surfaces of the cross-attention maps. Therefore, we set $\lambda^+$ equal for all queries inside the mask. Finally, (iii \& iv) we removed the positive and negative biases ($b^+$ and $b^-$; refer to \cref{eq:biases}). This is equivalent to removing the positive or negative influences of \cref{eq:score}.

We present the ablation results in \cref{tab:ablations} along with some visualization in \cref{fig:ablation} (see our \href{https://gato-vid.github.io/}{webpage} for full videos). The performance differences indicate that each component contributes positively to the localization metrics. Despite this, we observe some qualitative trade-offs. Ablating the projection (i) and Gaussian (ii) mechanisms has the least effect on the localization metrics, but has a higher impact on the SR and VBench metrics. The reduction in DD, AQ, and IQ suggests that excluding these components negatively affects dynamics and visuals. Next, the positive and negative components of \cref{eq:score} have the most significant impact. These two ablations demonstrate the most pronounced impact: a significant improvement in alignment comes at the expense of lower quality. In addition, and counterintuitively, removing the negative bias yielded the most severe degradation in localization performance.

\begin{figure}[t]
    \centering
    \begin{subfigure}[b]{0.32\linewidth}
        \includegraphics[width=\linewidth]{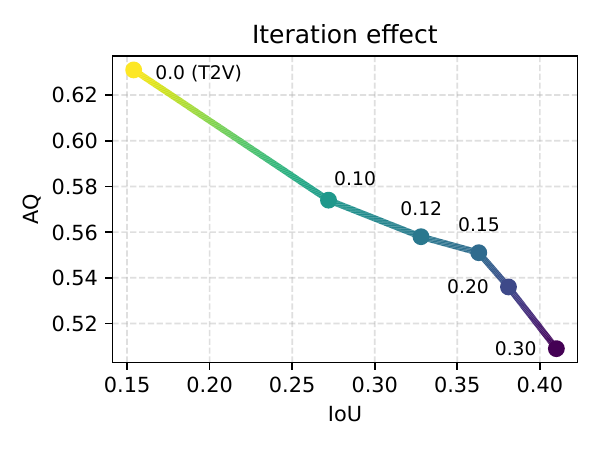}
        \caption{\textbf{Iteration ablation}}
        \label{fig:iters}
    \end{subfigure}
    \begin{subfigure}[b]{0.32\linewidth}
        \includegraphics[width=\linewidth]{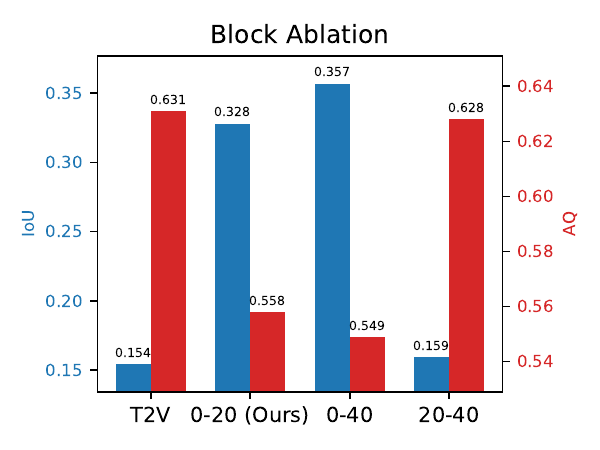}
        \caption{\textbf{Block ablation}}
        \label{fig:blocks}
    \end{subfigure}
    \begin{subfigure}[b]{0.32\linewidth}
        \includegraphics[width=\linewidth]{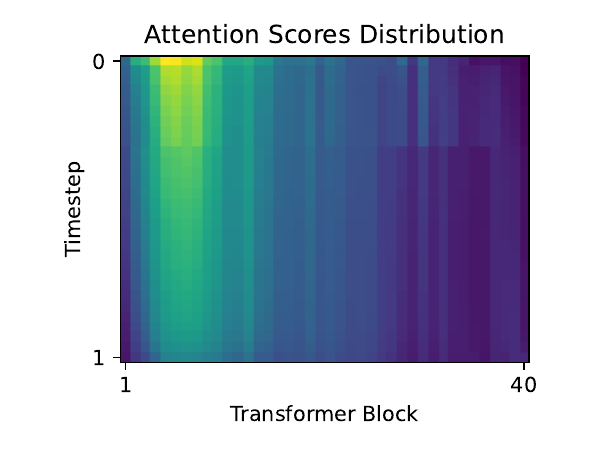}
        \caption{\textbf{Attention analysis}}
        \label{fig:attn}
    \end{subfigure}
    \caption{\textbf{(a):} More iterations induce higher IoU but reduce AQ. \textbf{(b):} Using the first 20 blocks is as effective as using all blocks, whereas using the last 20 blocks yields no improvement. \textbf{(c):} The early blocks produce higher attention scores than later blocks or steps, suggesting that initial layers specialize in the video's general layout.}
    \label{fig:iter-blocks}
    \vspace{-3mm}
\end{figure}


Finally, we analyze the effect of varying the injection scope across denoising steps and transformer blocks (\cref{fig:iter-blocks}). As shown in \cref{fig:iters}, applying our mechanism across a larger fraction of sampling iterations increases the IoU but gradually degrades AQ. Applying bias injection during the first 15\% of iterations strikes an optimal trade-off between spatial control and visual quality.
Regarding layer depth, injecting bias into the first 20 blocks achieves an IoU and AQ performance similar to applying it across all 40 blocks, whereas targeting only the last 20 blocks yields results close to the Wan2.2 (\cref{fig:blocks}). This behavior is explained by the cross-attention score distribution in \cref{fig:attn}, where early blocks show strong attention values, while later layers exhibit homogeneous scores, suggesting that late layers are less critical for structural layout.

\section{Conclusions}

We present \METHODNAME{}, a training- and gradient-free approach for spatially grounded text-to-video generation. By reformulating attention localization as an analytically solvable surrogate objective, our method replaces backpropagation-based approaches with a closed-form query steering mechanism. While our experiments suggest that while stronger spatial control can reduce scene dynamics and aesthetic appeal in some cases, they highlight the potential of analytical attention manipulation as an efficient alternative for controlling DiT-based architectures. We hope this work motivates future research into more accurate and scalable controllable video generation methods that better balance precise spatial control with visual fidelity.


\noindent \textbf{Acknowledgements} This work has been supported by chair VISA DEEP (ANR-20-CHIA-0022), Cluster PostGenAI@Paris
(ANR-23-IACL-0007, FRANCE 2030), and funded by the French National Research Agency (ANR) under the Renaissance project (grant ANR-23-CE23-0023; France 2030).
This work was performed using HPC resources from GENCI-IDRIS (Grant 2025-AD011017053).

%
%
\bibliographystyle{splncs04}
\bibliography{main}
\end{document}